\pdfoutput=1

\documentclass[11pt]{article}
\usepackage{hyperref}

\usepackage[]{ACL2023}

\usepackage{times}
\usepackage{latexsym}
\usepackage[T1]{fontenc}
\usepackage[utf8]{inputenc}

\usepackage{microtype}

\usepackage{inconsolata}
\usepackage{amsmath}
\usepackage{amsfonts}
\usepackage{subcaption}
\usepackage{graphicx}

\title{SmolTulu: Higher Learning Rate to Batch Size Ratios\\Can Lead to Better Reasoning in SLMs}

\author{
    Sultan Alrashed \\
    Saudi Data \& Artificial Intelligence Authority \\
    \texttt{srashed@sdaia.gov.sa}
}

\begin{document}
    \maketitle
    \begin{abstract}

        We present SmolTulu-1.7b-Instruct\footnote{Huggingface link:\url{https://huggingface.co/SultanR/SmolTulu-1.7b-Instruct}}, referenced in this report as SmolTulu-DPO-1130, an instruction-tuned language model that adapts AllenAI's Tulu 3 post-training pipeline \citep{lambert2024tulu3pushingfrontiers} to enhance Huggingface's SmolLM2-1.7B base model. Through comprehensive empirical analysis using a 135M parameter model, we demonstrate that the relationship between learning rate and batch size significantly impacts model performance in a task-dependent manner. Our findings reveal a clear split: reasoning tasks like ARC and GSM8K benefit from higher learning rate to batch size ratios, while pattern recognition tasks such as HellaSwag and IFEval show optimal performance with lower ratios. These insights informed the development of SmolTulu, which achieves state-of-the-art performance among sub-2B parameter models on instruction following, scoring 67.7\% on IFEval ($\Delta$11\%), and mathematical reasoning with 51.6\% on GSM8K ($\Delta$3.4\%), with an alternate version achieving scoring 57.1\% on ARC ($\Delta5.4\%$). We release our model, training recipes, and ablation studies to facilitate further research in efficient model alignment, demonstrating that careful adaptation of optimization dynamics can help bridge the capability gap between small and large language models.
    
    \end{abstract}

    \section{Introduction}
        Recent advances in language model post-training have demonstrated remarkable improvements in model capabilities through careful application of supervised finetuning (SFT), preference optimization, and reinforcement learning \citep{ouyang2022traininglanguagemodelsfollow, touvron2023llama2openfoundation}. However, these techniques have primarily been developed and validated on large language models with tens or hundreds of billions of parameters with smaller models being underexplored. The recently released Tulu 3 pipeline provides a comprehensive, open-source approach to post-training, but its effectiveness on significantly smaller models remains unexplored.
        
        Understanding how post-training dynamics scale to smaller models is crucial for democratizing access to high-quality language models and enabling deployment in resource-constrained environments. While \citep{shallue2019measuringeffectsdataparallelism} demonstrated that the relationship between batch size and training steps follows characteristic patterns across model families, they also found that the maximum useful batch size varies between workloads and depends heavily on model properties. This suggests that smaller models may require fundamentally different optimization strategies than their larger counterparts, especially across different downstream tasks.
    
        Through comprehensive ablations using Huggingface's SmolLM2-135M \citep{allal2024SmolLM2} as a base, we demonstrate that the relationship between learning rate and batch size significantly impacts model performance in a task-dependent manner during supervised finetuning. Our findings reveal a clear difference between reasoning and pattern recognition tasks: benchmarks requiring complex reasoning like ARC and GSM8K show optimal performance with higher learning rate to batch size ratios, while pattern recognition tasks such as HellaSwag and IFEval benefit from lower ratios. This aligns with theoretical work by \citep{masters2018revisitingsmallbatchtraining}, who argue that smaller batch sizes, despite their higher gradient variance, enable more frequent parameter updates that can better track the underlying objective's local structure.
    
        Furthermore, \citep{keskar2017largebatchtrainingdeeplearning} demonstrate that optimization trajectories can lead to qualitatively different solutions with varying generalization properties. Our empirical results suggest that higher learning rate to batch size ratios may help smaller models find flatter minima that generalize better to complex reasoning tasks, while lower ratios appear optimal for tasks dominated by pattern recognition. This provides new insight into how careful tuning of optimization parameters might help compensate for limited model capacity in a task-dependent manner.
    
        In this work, we investigate the adaptation of the Tulu 3 post-training pipeline \citep{lambert2024tulu3pushingfrontiers} to enhance SmolLM2-1.7B \citep{allal2024SmolLM2}, a compact language model with just 1.7 billion parameters. Through a set of ablations, we demonstrate that the relationship between learning rate and batch size plays a crucial role in determining model performance across different types of tasks.
    
        Our key contributions include:
        \begin{itemize}
            \item Empirical evidence demonstrating how learning rate to batch size ratios differentially affect reasoning and pattern recognition capabilities in small language models
            \item Release of SmolTulu, achieving state-of-the-art performance among sub-2B parameter models on instruction following and mathematical reasoning with verifiably no contamination.
            \item Theoretical analysis connecting optimization dynamics to task-specific performance in small language models
        \end{itemize}
    
        Our results indicate that while the core techniques from Tulu 3 translate effectively to smaller scales, achieving optimal performance requires careful consideration of how optimization dynamics change with model size and task type. Through these insights, we hope to contribute to the development of more efficient and accessible language models and establish new best practices for training smaller models that diverge from conventional wisdom derived from large-scale training with the learning rate linearly adjusted to the batch size.
        
    \section{Related Works}

        \subsection{Post-Training Recipes and Instruction Tuning}
            The development of modern post-training pipelines began with InstructGPT \citep{ouyang2022traininglanguagemodelsfollow}, which established the core supervised finetuning (SFT) and reinforcement learning from human feedback (RLHF) workflow. This approach has been widely adopted and adapted in open-source efforts, notably through works like Alpaca \citep{alpaca} and Vicuna \citep{vicuna2023}, which demonstrated the viability of instruction tuning with synthetic and user-generated data respectively. More recent work has focused on improving these pipelines, with projects like Tulu \citep{ivison2023camelschangingclimateenhancing} and Zephyr-$\beta$ \citep{tunstall2023zephyrdirectdistillationlm}, culminating in comprehensive open pipelines like Tulu 3 \citep{lambert2024tulu3pushingfrontiers}.

        \subsection{Direct Preference Optimization}
            Direct Preference Optimization (DPO) \citep{rafailov2024directpreferenceoptimizationlanguage} introduced a simplified approach to preference learning that eliminates the need for a separate reward model and complex RL training loops. This has been followed by various refinements and alternatives, including SLiC-HF \citep{zhao2023slichfsequencelikelihoodcalibration} and SimPO \citep{meng2024simpo}, which further simplified the training process while maintaining or improving performance. These developments have made preference learning more accessible and computationally efficient, particularly important for resource-constrained settings.
            
        \subsection{Verifiable Rewards and Reinforcement Learning}
            Recent work has explored using verifiable outcomes to improve model capabilities, particularly in domains like mathematics and coding. STaR \citep{zelikman2022starbootstrappingreasoningreasoning} pioneered the use of self-taught reasoning to improve mathematical capabilities, while TRICE \citep{phan2023trainingchainofthoughtlatentvariableinference} developed alternative approaches to learning from verifiable solutions. VinePPO \citep{kazemnejad2024vineppounlockingrlpotential} specifically targeted mathematical reasoning through reinforcement learning, demonstrating the effectiveness of RL approaches when ground truth answers are available. This line of work has shown particular promise in improving specific capabilities while maintaining model generality. Reinforcement Learning with Verifiable Rewards (RLVR) extends traditional RL approaches by using deterministic binary rewards based on ground truth outcomes rather than learned reward models\citep{lambert2024tulu3pushingfrontiers}. This approach has shown particular promise in domains with verifiable solutions like mathematics and programming, where correct outputs can be automatically validated.
            
        \subsection{Learning Rate and Batch Size Studies}
            The relationship between learning rate and batch size has been a subject of extensive study in deep learning. Foundational work by \citep{goyal2018accuratelargeminibatchsgd} established key principles for large-batch training, demonstrating that learning rate should be scaled proportionally with batch size to maintain training stability. \citep{smith2018dontdecaylearningrate} later proposed the counterintuitive strategy of increasing batch size instead of decaying learning rate, showing that this approach could achieve similar convergence while simplifying hyperparameter tuning.
            
            More recent work has focused on understanding these relationships in the context of transformer architectures. \citep{kaplan2020scalinglawsneurallanguage} provided empirical evidence that optimal training dynamics vary with model scale, while \citep{mccandlish2018empiricalmodellargebatchtraining} developed theoretical frameworks for understanding how batch size affects training efficiency. These insights have been particularly relevant for language models, where the interaction between learning rate, batch size, and model capacity can significantly impact the model's ability to learn different types of capabilities \citep{hoffmann2022trainingcomputeoptimallargelanguage}. However, most of this work has focused on large models, leaving open questions about how these relationships manifest in smaller architectures.

        \subsection{Efficient and Small Language Models}
            Research on efficient language models has seen renewed interest with the success of smaller but capable architectures. Notable examples include Microsoft's Phi models \citep{gunasekar2023textbooksneed}, which achieved strong reasoning capabilities at 1.3B parameters (50.6\% on HumanEval), and TinyLlama \citep{zhang2024tinyllamaopensourcesmalllanguage}, which adapted the Llama architecture to 1.1B parameters. The SmolLM family \citep{allal2024SmolLM2} represents a significant advance in this direction, demonstrating that carefully trained compact models can achieve competitive performance on a wide range of tasks while remaining deployable on consumer hardware.

    \section{Supervised Finetuning}
        \subsection{Dataset}
            To ensure fair evaluation, we analyzed the contamination levels of various benchmarks in the SFT dataset (allenai/tulu-3-sft-mixture). As shown in \autoref{tab:sft-contam}, most benchmarks exhibit minimal contamination rates below 1.5\%, with many showing zero contamination. The highest contamination rate was observed in PopQA at 7.21\%, while critical evaluation benchmarks like GSM8K, IFEval, and AGI Eval showed negligible to zero contamination, ensuring reliable performance measurements.

            \begin{table}[h]
                \centering
                \begin{tabular}{lc}
                    \hline
                    \textbf{Benchmark} & \textbf{Contamination} \\
                    \hline
                    cais/mmlu & 1.34\% \\
                    openai/openai\_humaneval & 0.00\% \\
                    openai/gsm8k & 0.08\% \\
                    ucinlp/drop & 0.20\% \\
                    lighteval/MATH & 0.06\% \\
                    google/IFEval & 0.00\% \\
                    akariasai/PopQA & 7.21\% \\
                    tatsu-lab/alpaca\_eval & 1.37\% \\
                    lukaemon/bbh & 0.02\% \\
                    truthfulqa/truthful\_qa & 1.47\% \\
                    allenai/wildguardmix & 0.06\% \\
                    allenai/wildjailbreak & 0.00\% \\
                    TIGER-Lab/MMLU-Pro & 0.93\% \\
                    Idavidrein/gpqa & 0.00\% \\
                    lighteval/agi\_eval\_en & 0.00\% \\
                    bigcode/bigcodebench & 0.00\% \\
                    deepmind/math\_dataset & 0.00\% \\
                    \hline
                \end{tabular}
                \caption{Contamination of benchmarks in the SFT dataset used \textit{allenai/tulu-3-sft-mixture}}
                \label{tab:sft-contam}
            \end{table}

        \subsection{Training}
        
        Foundation models have demonstrated remarkable capabilities across various tasks, yet their performance can be significantly influenced by the choice of hyperparameters during SFT. While work by \citep{smith2018dontdecaylearningrate} established foundational principles for batch size scaling in neural networks, and \citep{mccandlish2018empiricalmodellargebatchtraining} provided theoretical frameworks for understanding large-batch training, recent work by \citep{masters2018revisitingsmallbatchtraining} suggests that the conventional wisdom about batch size scaling may need reconsideration, particularly for smaller models.
        
                \begin{table*}[h]
                \small
                \centering
                \setlength{\tabcolsep}{4pt}
                \begin{tabular}{lcccc}
                    \hline
                    \textbf{Hyperparameter} & \textbf{SmolTulu} & \textbf{SmolTulu} & \textbf{Tulu 3} & \textbf{Tulu 3} \\
                    & \textbf{SFT-1130} & \textbf{SFT-1207} & \textbf{SFT 8b} & \textbf{SFT 70b} \\
                    \hline
                    Learning Rate (LR) & $9.0 \times 10^{-5}$ & $3.1 \times 10^{-6}$ & $5.0 \times 10^{-6}$ & $2.0 \times 10^{-6}$ \\
                    Batch Size (BS) & 8 & 32 & 128 & 128 \\
                    $\dfrac{LR}{BS} \times 10^{6}$ & 11.25 & 0.097 & 0.039 & 0.016 \\
                    \hline
                \end{tabular}
                \caption{SFT hyperparameter selection}
                \label{tab:sft-hyperparameters}
            \end{table*}

        Our experiments with SmolTulu variants build upon hyperparameter configurations established by AllenAI's extensive ablation studies for Tulu 3. As shown in \autoref{tab:sft-hyperparameters}, there is a notable pattern in the learning rate to batch size ratio (LR/BS) across model scales: as models grow larger, this ratio tends to decrease. This aligns with the scaling laws observed by \citep{kaplan2020scalinglawsneurallanguage} and the compute-optimal training strategies described in \citep{hoffmann2022trainingcomputeoptimallargelanguage}. The Tulu 3 70B model uses the smallest ratio of \(0.016 \times 10^{-6}\), while our SmolTulu SFT-1130 variant employs a significantly larger ratio of \(11.25 \times 10^{-6}\).

        \subsection{Results \& Discussion}
        
            \begin{table}[h]
                \small
                \centering
                \setlength{\tabcolsep}{4pt}
                \begin{tabular}{lccc}
                    \hline
                    \textbf{Metric} & \textbf{SmolTulu} & \textbf{SmolTulu} & \textbf{SmolLM2} \\
                    & \textbf{SFT-1130} & \textbf{SFT-1207} & \textbf{1.7B-Instruct} \\
                    \hline
                    ARC (Average) & 51.0 & \textbf{55.6} & 51.7 \\
                    BBH (3-shot) & \textbf{34.7} & 34.0 & 32.2 \\
                    GSM8K (5-shot) & \textbf{49.0} & 42.8 & 48.2 \\
                    HellaSwag & 61.5 & \textbf{67.5} & 66.1 \\
                    IFEval (Average) & \textbf{61.0} & 47.8 & 56.7 \\
                    MMLU-Pro (MCF) & 17.6 & 17.9 & \textbf{19.3} \\
                    PIQA & 72.7 & \textbf{76.9} & 74.4 \\
                    \hline
                \end{tabular}
                \caption{Performance comparison of SFT models}
                \label{tab:sft-comparison}
            \end{table}

        \begin{figure*}[t]
            \centering
            \begin{subfigure}[t]{0.48\textwidth}
                \centering
                \includegraphics[width=\textwidth]{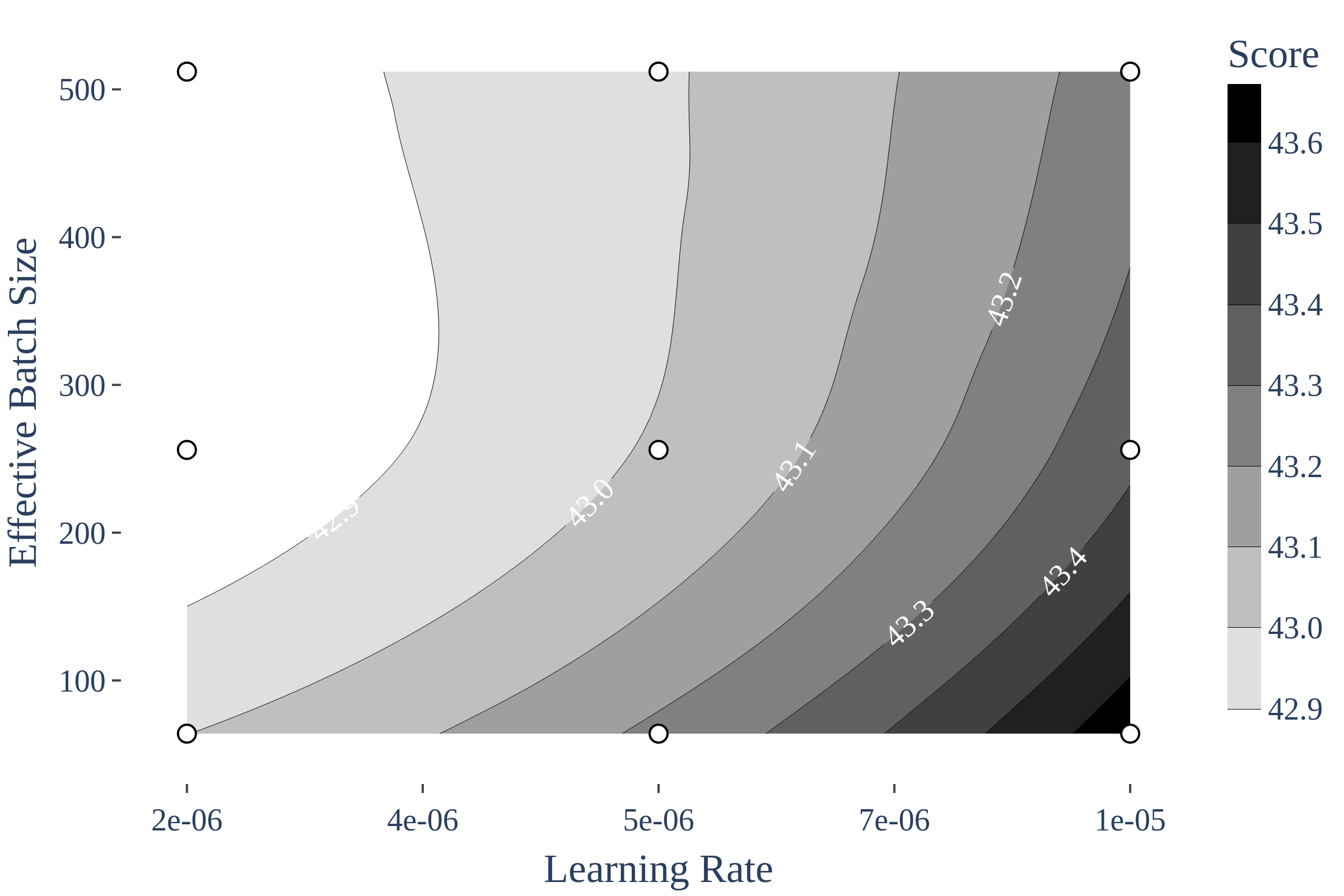}
                \caption{Effect of learning rate and batch size on ARC score.}
                \label{fig:arc_contour}
            \end{subfigure}%
            \hfill
            \begin{subfigure}[t]{0.48\textwidth}
                \centering
                \includegraphics[width=\textwidth]{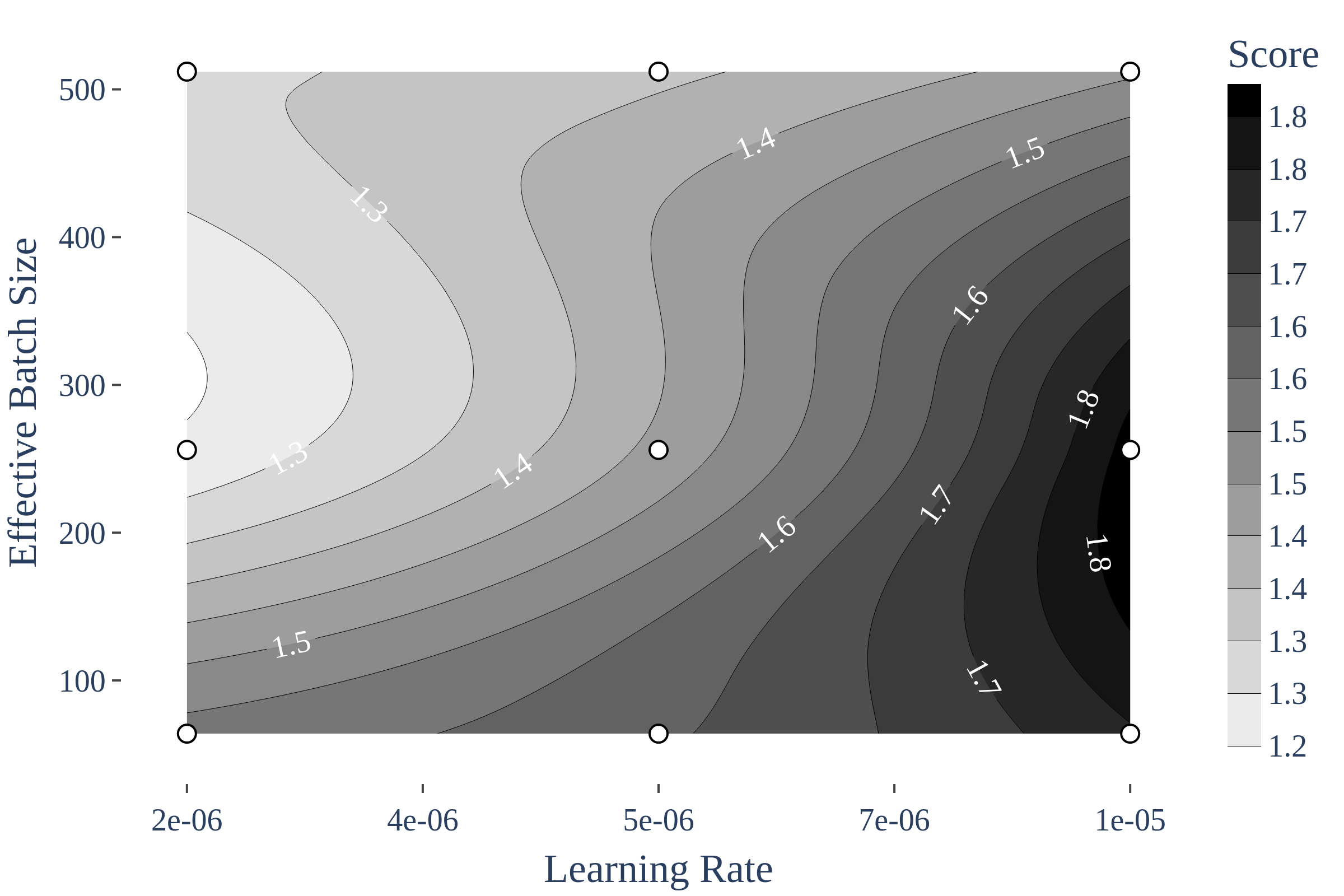}
                \caption{Effect of learning rate and batch size on GSM8K score.}
                \label{fig:gsm8k_contour}
            \end{subfigure}
            
            \vspace{1em}
            
            \begin{subfigure}[t]{0.48\textwidth}
                \centering
                \includegraphics[width=\textwidth]{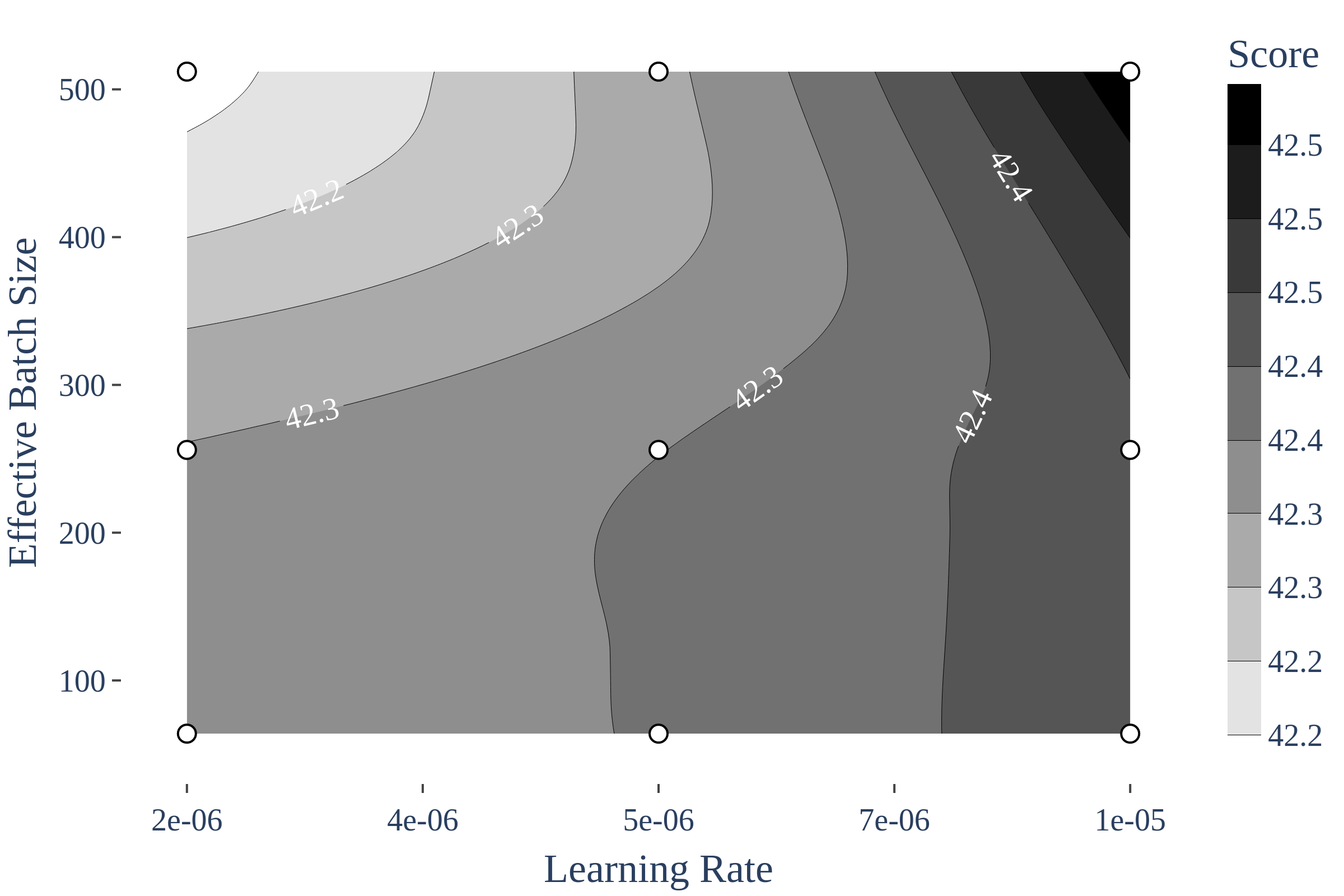}
                \caption{Effect of learning rate and batch size on HellaSwag score.}
                \label{fig:hellaswag_contour}
            \end{subfigure}%
            \hfill
            \begin{subfigure}[t]{0.48\textwidth}
                \centering
                \includegraphics[width=\textwidth]{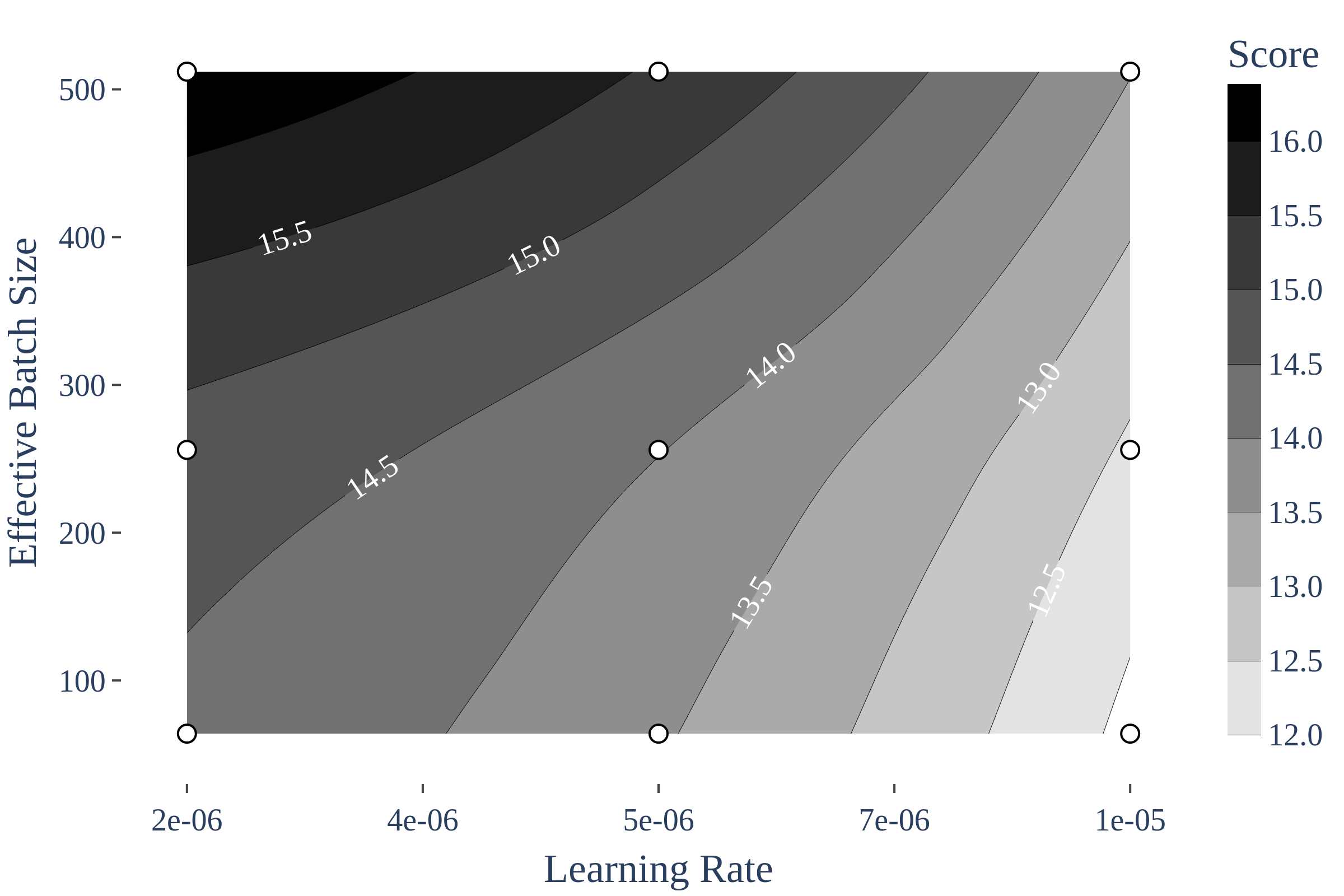}
                \caption{Effect of learning rate and batch size on IFEval score.}
                \label{fig:ifeval_contour}
            \end{subfigure}
            \caption{Contour analysis of learning rate and batch size effects on different evaluation metrics during supervised finetuning of SmolLM2-135M. The color scales represent scores for each metric, with black indicating higher performance. The patterns reveal task-dependent optimal ratios between learning rate and batch size.}
            \label{fig:contours}
        \end{figure*}

        The supervised finetuning experiments reveal complex relationships between optimization dynamics and model capabilities that evolve with model scale. Our initial ablation study using a 135M parameter model demonstrates clear task-dependent patterns in the relationship between learning rate and batch size ratios.

        As shown in \autoref{fig:arc_contour} and \autoref{fig:gsm8k_contour}, reasoning tasks like ARC and GSM8K consistently benefit from higher learning rate to batch size ratios. This aligns with theoretical work by \citep{keskar2017largebatchtrainingdeeplearning}. The improvement is particularly pronounced for GSM8K, where performance increases monotonically with the learning rate to batch size ratio.
    
        Conversely, at the 135M scale, pattern recognition tasks show markedly different behavior. \autoref{fig:hellaswag_contour} and \autoref{fig:ifeval_contour} reveal that HellaSwag and IFEval achieve optimal performance with lower learning rate to batch size ratios. This contrast suggests that at smaller scales, different types of learning may benefit from fundamentally different optimization dynamics, supporting \citep{shallue2019measuringeffectsdataparallelism}'s observation that optimal batch sizes vary significantly between workloads.
    
        However, these relationships become more nuanced at larger model scales. In our 1.7B parameter model, while GSM8K continues to benefit from higher ratios (achieving 51.6\% with SmolTulu-1130's high ratio versus 44.7\% with SmolTulu-1207's lower ratio), the pattern for other tasks shifts. Notably, IFEval performance improves with higher ratios (67.7\% vs 56.6\%), while ARC and MMLU-Pro show optimal performance with lower ratios (57.1\% and 19.1\% respectively).
    
        This scale-dependent shift in optimization dynamics aligns with \citep{masters2018revisitingsmallbatchtraining}'s argument that the relationship between batch size and learning rate fundamentally changes with model capacity. We hypothesize that at 135M parameters, the limited model capacity forces a strict trade-off between different types of learning, requiring distinct optimization strategies for reasoning versus pattern recognition. At 1.7B parameters, the increased capacity appears to enable more flexible learning dynamics, allowing both GSM8K and IFEval to benefit from aggressive optimization while tasks like ARC and MMLU-Pro benefit from more conservative approaches.
    
        These findings suggest that the conventional wisdom about learning rate and batch size relationships derived from large-scale training may need significant adaptation for smaller models. The interaction between model capacity and optimal optimization strategy appears more complex previously thought.
            
        \section{Direct Preference Optimization}
    
        \subsection{Dataset}
            We conducted contamination analysis on the DPO dataset (allenai/llama-3.1-tulu-3-8b-preference-mixture) to verify evaluation fairness. \autoref{tab:dpo-contam} shows consistently low contamination rates across benchmarks, with most falling below 1\%. The highest contamination was found in PopQA at 2.72\%, while crucial benchmarks like GSM8K, IFEval, and BBH maintained zero contamination, enabling trustworthy assessment of model improvements through preference optimization.
    
            \begin{table}[h]
                \centering
                \begin{tabular}{lc}
                    \hline
                    \textbf{Benchmark} & \textbf{Contamination} \\
                    \hline
                    cais/mmlu & 0.69\% \\
                    openai/openai\_humaneval & 0.00\% \\
                    openai/gsm8k & 0.00\% \\
                    ucinlp/drop & 0.07\% \\
                    lighteval/MATH & 0.02\% \\
                    google/IFEval & 0.00\% \\
                    akariasai/PopQA & 2.72\% \\
                    tatsu-lab/alpaca\_eval & 1.24\% \\
                    lukaemon/bbh & 0.00\% \\
                    truthfulqa/truthful\_qa & 0.61\% \\
                    allenai/wildguardmix & 0.06\% \\
                    allenai/wildjailbreak & 0.00\% \\
                    TIGER-Lab/MMLU-Pro & 0.36\% \\
                    Idavidrein/gpqa & 0.00\% \\
                    lighteval/agi\_eval\_en & 0.00\% \\
                    bigcode/bigcodebench & 0.00\% \\
                    deepmind/math\_dataset & 0.00\% \\
                    \hline
                \end{tabular}
                \caption{Contamination of benchmarks in the DPO dataset used \textit{allenai/llama-3.1-tulu-3-8b-preference-mixture}}
                \label{tab:dpo-contam}
            \end{table}
    
        \subsection{Training}

             We employed Direct Preference Optimization (DPO) for preference learning while adapting the hyperparameters for our smaller model scale, using the same pipeline as described in Tulu 3 \citep{lambert2024tulu3pushingfrontiers}. DPO allows direct optimization of the policy without requiring a separate reward model \citep{rafailov2024directpreferenceoptimizationlanguage}. The core objective function is shown in equation \ref{eq:dpo}.
        
            \begin{figure*}[t]
                \begin{equation}\label{eq:dpo}
                    \max_{\pi_\theta} \mathbb{E}_{y_c,y_r\sim\mathcal{D}}\left[\log \sigma\left(\beta \log \frac{\pi_\theta(y_c|x)}{\pi_{\text{ref}}(y_c|x)} - \beta \log \frac{\pi_\theta(y_r|x)}{\pi_{\text{ref}}(y_r|x)}\right)\right]
                \end{equation}
                \caption{Direct Preference Optimization objective function, where $\pi_{\text{ref}}$ is the reference model (our SFT model) and $\beta$ controls KL divergence from the reference model.}
            \end{figure*}
            
            We used the length-normalized variant of DPO, which showed superior performance in Tulu 3's ablations, where the log probabilities are normalized by sequence length.
                    
            \begin{itemize}
                \item Pre-computing and caching log probabilities from the reference model instead of keeping it in memory \citep{hu2024openrlhfeasytousescalablehighperformance}
                \item Performing separate forward passes for chosen and rejected sequences rather than concatenating them
            \end{itemize}
    
            \begin{table*}[h]
                \small
                \centering
                \setlength{\tabcolsep}{4pt}
                \begin{tabular}{lcccc}
                    \hline
                    \textbf{Hyperparameter} & \textbf{SmolTulu} & \textbf{SmolTulu} & \textbf{Tulu 3} & \textbf{Tulu 3} \\
                    & \textbf{DPO-1130} & \textbf{DPO-1207} & \textbf{DPO 8b} & \textbf{DPO 70b} \\
                    \hline
                    Learning Rate (LR) & $8.0 \times 10^{-7}$ & $5 \times 10^{-7}$ & $5.0 \times 10^{-7}$ & $2.0 \times 10^{-7}$ \\
                    Batch Size (BS) & 12 & 32 & 128 & 128 \\
                    $\dfrac{LR}{BS} \times 10^{7}$ & 0.667 & 0.156 & 0.039 & 0.016 \\
                    \hline
                \end{tabular}
                \caption{DPO hyperparameter selection}
                \label{tab:dpo-hyperparameters}
            \end{table*}
    
            However, our experiments revealed that smaller models benefit from different hyperparameter configurations than their larger counterparts. As shown in \autoref{tab:dpo-hyperparameters}, we maintained higher learning rate to batch size ratios compared to Tulu 3's models, with SmolTulu DPO-1130 using a ratio approximately 17 times larger than Tulu 3 8B's.
    
            For DPO-1130, we used a learning rate of $8.0 \times 10^{-7}$ and batch size of 12, while DPO-1207 used a more conservative learning rate of $5.0 \times 10^{-7}$ with batch size 32. Both variants used:
    
            \begin{itemize}
                \item Maximum sequence length of 2,048 tokens
                \item KL penalty coefficient $\beta = 5$
                \item Linear learning rate schedule with 0.1 warmup ratio
                \item Single training epoch
            \end{itemize}
    
            Our experimentation with higher learning rate to batch size ratios was motivated by the hypothesis that smaller models may require larger per-example updates to effectively learn from preference data, particularly for complex reasoning tasks where they lack the inherent capacity of larger models.

        \subsection{Results \& Discussion}
        
            \begin{table}[h]
                \small
                \centering
                \setlength{\tabcolsep}{4pt}
                \begin{tabular}{lccc}
                    \hline
                    \textbf{Metric} & \textbf{SmolTulu} & \textbf{SmolTulu} & \textbf{SmolLM2} \\
                    & \textbf{DPO-1130} & \textbf{DPO-1207} & \textbf{1.7B-Instruct} \\
                    \hline
                    ARC (Average) & 51.5 & \textbf{57.1} & 51.7 \\
                    BBH (3-shot) & \textbf{33.8} & \textbf{34.2} & 32.2 \\
                    GSM8K (5-shot) & \textbf{51.6} & 44.7 & 48.2 \\
                    HellaSwag & 61.1 & 64.2 & \textbf{66.1} \\
                    IFEval (Average) & \textbf{67.7} & 56.6 & 56.7 \\
                    MMLU-Pro (MCF) & 17.4 & 19.1 & \textbf{19.3} \\
                    PIQA & 72.2 & \textbf{76.4} & 74.4 \\
                    \hline
                \end{tabular}
                \caption{Performance comparison of DPO models}
                \label{tab:dpo-comparison}
            \end{table}
    
            The DPO experiments yielded notable improvements over the SFT baseline across several key metrics. As shown in \autoref{tab:dpo-comparison}, SmolTulu DPO-1130, using the higher learning rate to batch size ratio, achieved substantial gains on instruction following and mathematical reasoning tasks, reaching 67.7\% on IFEval and 51.6\% on GSM8K. Conversely, SmolTulu DPO-1207, employing a lower learning rate to batch size ratio more similar to larger models, showed stronger performance on pattern recognition tasks like ARC (57.1\%) and PIQA (76.4\%).
        
            However, our exploration of hyperparameter configurations was significantly constrained by computational resources. Given that DPO builds directly upon the SFT model, the initial policy's convergence properties may significantly influence subsequent preference learning. A comprehensive understanding would require extensive ablation studies across different SFT checkpoints.
            
    \section{Reward Modelling}
    
        \subsection{Dataset}
            Our reward models (RM) were trained on the same preference dataset used in the DPO stage, combining UltraFeedback with additional synthetic preference data generated through the Tulu 3 pipeline. We maintained rigorous contamination controls across all evaluation benchmarks, as detailed in earlier sections.
    
        \subsection{Training}
            \begin{table*}[h]
                \small
                \centering
                \setlength{\tabcolsep}{4pt}
                \begin{tabular}{lccc}
                    \hline
                    \textbf{Hyperparameter} & \textbf{SmolTulu} & \textbf{SmolTulu} & \textbf{Tulu 3} \\
                    & \textbf{RM-1130} & \textbf{RM-1207} & \textbf{DPO 8b} \\
                    \hline
                    Learning Rate (LR) & $4.0 \times 10^{-5}$ & $7.5 \times 10^{-7}$ & $5.0 \times 10^{-7}$ \\
                    Batch Size (BS) & 4 & 8 & 128 \\
                    $\dfrac{LR}{BS} \times 10^{7}$ & 100 & 0.938 & 0.039 \\
                    \hline
                \end{tabular}
                \caption{Reward model hyperparameter selection}
                \label{tab:rm-hyperparameters}
            \end{table*}
    
            Following the Tulu 3 methodology \citep{lambert2024tulu3pushingfrontiers}, we trained our reward models using the standard pairwise preference learning objective. Given a preference dataset $\mathcal{D}$ consisting of prompts x and two responses per prompt $(y, y')$, where one response is chosen \(y_{c}\) and one is rejected \(y_{r}\), the reward model \(r_{\phi}\) is trained to maximize:
    
            \begin{equation}
                \max_{r_\phi} \mathbb{E}_{(x,y_c,y_r)\sim\mathcal{D}}[\log \sigma(r_\phi(x, y_c) - r_\phi(x, y_r))]
            \end{equation}
    
            where $\sigma$ is the logistic function. This objective maximizes the difference between rewards, with this difference representing the log-likelihood that \(y_{c}\) will be preferred over \(y_{r}\). We used the same several key implementation details from Tulu 3, except for ones that involved batch size and learning rate, where our changes can be seen in \autoref{tab:rm-hyperparameters}.
    
        \subsection{Results \& Discussion}
            \begin{table}[h]
                \small
                \centering
                \setlength{\tabcolsep}{4pt}
                \begin{tabular}{lcccc}
                    \hline
                    \textbf{Metric} & \textbf{SmolTulu} & \textbf{SmolTulu} & \textbf{Tulu 3} \\
                    & \textbf{RM-1130} & \textbf{RM-1207} & \textbf{8b RM} \\
                    \hline
                    RB Chat & \textit{94.13} & 83.52 & \textbf{96.27} \\
                    RB Chat Hard & 43.64 & \textit{44.74} & \textbf{55.92} \\
                    RB Safety & \textit{75.54} & 64.59 & \textbf{84.05} \\
                    RB Reasoning & \textit{68.01} & 54.71 & \textbf{76.50} \\
                    RB Average & \textit{72.43} & 58.59 & \textbf{81.34} \\
                    UFB & \textit{73.17} & 61.66 & \textbf{77.34} \\
                    \hline
                \end{tabular}
                \caption{Performance comparison of reward models, where UFB is the \textit{test\_prefs} split of \textit{allenai/ultrafeedback\_binarized\_cleaned} and RB is RewardBench.}
                \label{tab:reward-comparison}
            \end{table}

        The reward modeling experiments revealed interesting patterns consistent with our findings from SFT and DPO stages. SmolTulu RM-1130, employing a much larger learning rate to batch size ratio, demonstrated strong performance across various metrics on RewardBench (RB) \citep{lambert2024rewardbench}, achieving 94.13\% on standard chat evaluation and 75.54\% on safety assessments, as seen in \autoref{tab:reward-comparison}. This pattern of strong relative performance extends across other metrics, with SmolTulu RM-1130 achieving 73.17\% accuracy on UltraFeedback benchmark test preferences, falling only 4.17 percentage points short of Tulu 3's 77.34\% despite using approximately 21\% of the parameters. Following \citep{shallue2019measuringeffectsdataparallelism}'s framework, these results suggest that reward modeling may scale more gracefully to smaller architectures than previously assumed, particularly when using appropriately adapted optimization strategies.

        The substantial performance gap between RM-1130 and RM-1207 (72.43\% vs 58.59\% on RB reinforces our earlier findings about the importance of learning rate to batch size ratios in smaller models. The higher ratio used in RM-1130 appears to be particularly crucial for reward modeling, where the task of learning preference relationships may benefit from larger per-example updates and more frequent gradient computations. However, establishing the precise nature of this relationship would require more extensive ablation studies, which we leave to future work with greater computational resources.

    \section{Reinforcement Learning with Verifiable Rewards}

            While our initial experiments with RLVR showed promise, the computational requirements for thorough hyperparameter exploration proved prohibitive.

            Our preliminary investigations suggest that the relationship between learning rate and batch size may be particularly complex in the RLVR setting, where the sparse binary reward signal introduces additional optimization challenges for smaller models. However, establishing concrete findings would require extensive ablation studies beyond our current computational resources. We leave comprehensive exploration of these dynamics to future work with greater computational capacity.
                        
    \section{Limitations}
        While our work demonstrates promising results in adapting the Tulu 3 pipeline to smaller models, several important limitations should be noted:
    
        \paragraph{Base Model Dependencies} Our findings are specific to SmolLM2 and may not generalize to other small models, particularly those with different pretraining approaches or architectural choices.
    
        \paragraph{Multi-Stage Optimization Understanding} The relationship between optimization choices in different training stages (SFT, DPO, RM, RLVR) remains poorly understood, especially in the context of smaller models. While we observed consistent benefits from higher learning rate to batch size ratios, the theoretical foundations for this interaction across training stages requires further investigation.
    
    \section{Conclusion}

        \begin{table*}[t]
        \small
        \centering
        \setlength{\tabcolsep}{4pt}
        \begin{tabular}{lccccccc}
        \hline
        \textbf{Metric} & \textbf{SmolTulu} & \textbf{SmolTulu} & \textbf{SmolTulu} & \textbf{SmolTulu} & \textbf{SmolLM2} & \textbf{Llama-3.2} & \textbf{Qwen2.5} \\
        & \textbf{DPO-1130} & \textbf{DPO-1207} & \textbf{SFT-1130} & \textbf{SFT-1207} & \textbf{1.7B-Instruct} & \textbf{1B-Instruct} & \textbf{1.5B-Instruct} \\
        \hline
        ARC (Average) & 51.5 & \textbf{57.1} & 51.0 & 55.6 & 51.7 & 41.6 & 46.2 \\
        BBH (3-shot) & 33.8 & 34.2 & 34.7 & 34.0 & 32.2 & 27.6 & \textbf{35.3}\\
        GSM8K (5-shot) & \textbf{51.6} & 44.7 & 49.0 & 42.8 & 48.2 & 26.8 & 42.8 \\
        HellaSwag & 61.1 & 64.2 & 61.5 & \textbf{67.5} & 66.1 & 56.1 & 60.9 \\
        IFEval (Average) & \textbf{67.7} & 56.6 & 61.0 & 47.8 & 56.7 & 53.5 & 47.4  \\
        MMLU-Pro (MCF) & 17.4 & 19.1 & 17.6 & 17.9 & 19.3 & 12.7 & \textbf{24.2} \\
        PIQA & 72.2 & 76.4 & 72.7 & \textbf{76.9} & 74.4 & 72.3 & 73.2 \\
        \hline
        \end{tabular}
        \caption{A comparison against a wider selection of models}
        \label{tab:full-comparison}
        \end{table*}
    
        We have demonstrated that careful adaptation of modern post-training techniques can yield strong results even at significantly smaller model scales. We found that smaller models may require substantially different optimization dynamics than their larger counterparts to achieve optimal performance. Our empirical results align with theoretical frameworks from optimization literature, suggesting that higher ratios can help compensate for limited model capacity, particularly on complex reasoning tasks.
    
        The resulting model, SmolTulu, achieves state-of-the-art performance among sub-2B parameter models on instruction following while maintaining strong mathematical reasoning capabilities as seen in \autoref{tab:full-comparison}. These results suggest that the effectiveness of post-training pipelines like Tulu 3 can extend to much smaller scales when properly adapted. Our findings indicate that optimal training strategies may need to vary significantly based on both model scale and target capabilities.
    
        Looking forward, we believe this work opens up promising directions for making high-quality language models more accessible and deployable in resource-constrained environments. Future work investigating adaptive optimization strategies that account for both model scale and task requirements could further advance this goal. Additionally, developing theoretical frameworks that specifically address the interaction between model capacity and optimization dynamics could help establish more principled approaches to training smaller models.
        
        \section{Acknowledgements}
        Thank you Faisal Alhejary, Abdulmajeed Alrowaithy, Tariq Aljaber, Abdulaziz Albuainain, and Salman Alsubaihi for providing me with their support. Thank you AllenAI and Huggingface for your continual contributions to open-source.
    
    \bibliography{main}
    \bibliographystyle{acl_natbib}

    \appendix

    \section{Language Distribution in Datasets}
        Using \textit{XLM-RoBERTa} for language detection, we can get an estimate for the presence of each language in a given dataset.
    
        \begin{table}[h]
            \centering
            \begin{tabular}{lc}
                \hline
                \textbf{Language} & \textbf{Presence (\%)} \\
                \hline
                English & 83.13 \\
                Hindi & 3.79 \\
                Swahili & 2.02 \\
                Russian & 2.00 \\
                Spanish & 1.15 \\
                Arabic & 0.98 \\
                Chinese & 0.94 \\
                Turkish & 0.87 \\
                Urdu & 0.78 \\
                Portuguese & 0.77 \\
                Vietnamese & 0.64 \\
                Japanese & 0.63 \\
                French & 0.66 \\
                Bulgarian & 0.33 \\
                Italian & 0.32 \\
                Dutch & 0.31 \\
                Polish & 0.25 \\
                German & 0.23 \\
                Thai & 0.10 \\
                Greek & 0.09 \\
                \hline
            \end{tabular}
            \caption{Language distribution in SFT dataset.}
            \label{tab:sft-language-distribution}
        \end{table}

        \begin{table}[h]
            \centering
            \begin{tabular}{lc}
                \hline
                \textbf{Language} & \textbf{Presence (\%)} \\
                \hline
                English & 86.24 \\
                Hindi & 2.23 \\
                Russian & 2.03 \\
                French & 1.42 \\
                Spanish & 1.40 \\
                Chinese & 1.37 \\
                Urdu & 0.68 \\
                Swahili & 0.65 \\
                German & 0.58 \\
                Japanese & 0.57 \\
                Portuguese & 0.54 \\
                Arabic & 0.51 \\
                Turkish & 0.42 \\
                Vietnamese & 0.33 \\
                Italian & 0.32 \\
                Polish & 0.22 \\
                Dutch & 0.18 \\
                Bulgarian & 0.18 \\
                Thai & 0.10 \\
                Greek & 0.04 \\
                \hline
            \end{tabular}
            \caption{Language distribution in DPO / RM dataset.}
            \label{tab:dpo-language-distribution}
        \end{table}

        \begin{table}[h]
            \centering
            \begin{tabular}{lc}
                \hline
                \textbf{Language} & \textbf{Presence (\%)} \\
                \hline
                English & 94.80 \\
                French & 1.29 \\
                Spanish & 1.04 \\
                Chinese & 0.66 \\
                German & 0.55 \\
                Russian & 0.48 \\
                Japanese & 0.40 \\
                Hindi & 0.23 \\
                Polish & 0.10 \\
                Portuguese & 0.10 \\
                Dutch & 0.08 \\
                Urdu & 0.07 \\
                Bulgarian & 0.07 \\
                Italian & 0.05 \\
                Turkish & 0.03 \\
                Arabic & 0.03 \\
                Vietnamese & 0.02 \\
                Swahili & 0.00 \\
                \hline
            \end{tabular}
            \caption{Language distribution in RLVR dataset.}
            \label{tab:rlvr-language-distribution}
        \end{table}

    \section{RLVR Dataset Contamination}

        We included the contamination results of our intended RLVR dataset for reference.

        \begin{table}[h]
            \centering
            \begin{tabular}{lc}
                \hline
                \textbf{Benchmark} & \textbf{Contamination} \\
                \hline
                cais/mmlu & 0.65\% \\
                openai/openai\_humaneval & 0.00\% \\
                openai/gsm8k & 0.00\% \\
                ucinlp/drop & 0.00\% \\
                lighteval/MATH & 0.24\% \\
                google/IFEval & 0.00\% \\
                akariasai/PopQA & 0.45\% \\
                tatsu-lab/alpaca\_eval & 0.12\% \\
                lukaemon/bbh & 0.00\% \\
                truthfulqa/truthful\_qa & 0.12\% \\
                allenai/wildguardmix & 0.00\% \\
                allenai/wildjailbreak & 0.00\% \\
                TIGER-Lab/MMLU-Pro & 0.66\% \\
                Idavidrein/gpqa & 0.00\% \\
                lighteval/agi\_eval\_en & 0.00\% \\
                bigcode/bigcodebench & 0.00\% \\
                deepmind/math\_dataset & 0.00\% \\
                \hline
            \end{tabular}
            \caption{Contamination of benchmarks in the RLVR dataset \textit{allenai/RLVR-GSM-MATH-IF-Mixed-Constraints}}
            \label{tab:rlvr-contam}
        \end{table}

\end{document}